%% file: main.tex
\documentclass[a4paper,fleqn]{cas-dc}

\usepackage[T1]{fontenc}
\usepackage[utf8]{inputenc}
\usepackage{microtype}
\usepackage{tabularx}
\usepackage{xspace}
\usepackage{amsmath}
\usepackage{graphicx}
\graphicspath{{./}}
\usepackage{booktabs}
\usepackage{multirow}
\usepackage{xcolor,colortbl}
\usepackage[numbers]{natbib}
\usepackage{float}

\definecolor{GrayBG}{gray}{0.95}
\definecolor{lightgreen}{RGB}{231,255,219}
\newcommand{\greentext}[1]{\colorbox{lightgreen}{#1}\xspace}
\def\highlightcolor{black}
\newcommand{\marktext}[1]{\textcolor{\highlightcolor}{#1}}

\begin{document}
\let\WriteBookmarks\relax
\def\floatpagepagefraction{1}
\def\textpagefraction{.001}

\title[mode = title]{KERMIT: Knowledge Graph Completion of Enhanced Relation Modeling with Inverse Transformation}
\shorttitle{KERMIT: Knowledge Graph Completion of Enhanced Relation Modeling with Inverse Transformation}

\shortauthors{H. Li et~al.}
\author[1]{Haotian Li}[orcid=0000-0002-4708-3953]
\ead{22b903069@stu.hit.edu.cn}
\fnmark[1]
\credit{Software, Data Curation, Writing – Original Draft}

\author[2]{Bin Yu}[orcid=0000-0002-2132-953X]
\ead{22s130455@stu.hit.edu.cn}
\fnmark[1]
\credit{Software, Data Curation, Writing – Original Draft}

\author[2,3]{Yuliang Wei}[orcid=0009-0004-4688-6393]
\ead{wei.yl@hit.edu.cn}
\credit{Validation, Writing – Review}

\author[2]{Kai Wang}[orcid=0000-0003-3044-9047]
\ead{dr.wangkai@hit.edu.cn}
\credit{Writing – Review}

\author[4]{Richard Yi Da Xu}[orcid=0000-0003-2080-4762]
\ead{xuyida@hkbu.edu.hk}
\credit{Conceptualization, Methodology, Writing – Review}
\fnmark[2]

\author[1]{Bailing Wang}[orcid=0000-0003-2973-8036]
\ead{wbl@hit.edu.cn}
\credit{Conceptualization, Funding acquisition}
\cormark[1]
\fnmark[2]

\affiliation[1]{
organization={Research Institute of Cyberspace Security, Harbin Institute of Technology},
city={Harbin},
country={China}
}

\affiliation[2]{
organization={School of Computer Science and Technology, Harbin Institute of Technology at Weihai},
city={Weihai},
country={China}
}

\affiliation[3]{
organization={Shandong Key Laboratory of Industrial Network Security},
state={Shandong},
country={China}
}

\affiliation[4]{
organization={Department of Mathematics, Hong Kong Baptist University},
city={Hongkong},
country={China}
}

\cortext[cor1]{Corresponding author}
\fntext[fn1]{Equal contributing}
\fntext[fn2]{Equal advising}

\begin{abstract}
Knowledge graph completion (KGC) revolves around populating missing triples in a knowledge graph using available information. Text-based methods, which depend on textual descriptions of triples, often encounter difficulties when these descriptions lack sufficient information for accurate prediction, an issue inherent to the datasets and not easily resolved through modeling alone. To address this and ensure data consistency, we first use large language models (LLMs) to generate coherent descriptions, bridging the semantic gap between queries and answers. Secondly, we utilize inverse relations to create a symmetric graph, thereby providing augmented training samples for KGC. Additionally, we employ the label information inherent in knowledge graphs (KGs) to enhance the existing contrastive framework, making it fully supervised. These efforts have led to significant performance improvements on the WN18RR, FB15k-237 and UMLS datasets. According to standard evaluation metrics, our approach achieves a 3.0\% improvement in Hit@1 on WN18RR and a 12.1\% improvement in Hit@3 on UMLS, demonstrating superior performance.
\end{abstract}

\begin{keywords}
Knowledge graph completion (KGC) \sep Large language models (LLMs) \sep Supervised contrastive learning
\end{keywords}

\maketitle

\input{introduction}
\input{related_work}
\input{preliminaries}
\input{methodology}
\input{experiments}
\input{analysis}
\input{conclusion}
\input{broader_impact}
\input{acknowledgements}

\printcredits

\bibliographystyle{elsarticle-num}
\bibliography{reference}

\clearpage
\appendix
\onecolumn
\input{appendix}

\end{document}

%% file: introduction.tex
\section{Introduction}
Knowledge graphs (KGs) have emerged as powerful tools for representing structured information, driving progress in areas like natural language processing, data mining, and artificial intelligence. KGs encode knowledge in triples $(h, r, t)$, where $h$ is the head entity, $t$ the tail entity, and $r$ the relationship, forming a rich semantic network. \marktext{Their applications span recommendation systems \cite{guo2020survey}, question answering \cite{fu2020survey}, social network analysis \cite{he2020constructing}, and cyber security \cite{jia2023artificial}}. However, real-world KGs (e.g., Freebase \cite{bollacker2008freebase}, DBpedia \cite{auer2007dbpedia}, YAGO \cite{suchanek2007yago}) often suffer from incompleteness, missing vital relationships and entities \cite{dong2014knowledge}. This issue drives the need for knowledge graph completion (KGC), which aims to fill these gaps by inferring missing information and enhancing the graph's utility. Typically, a KGC model seeks to identify the most plausible tail entity $t$ for a given query $(h, r, ?)$.

\begin{figure*}
  \centering
  \includegraphics[width=1\textwidth]{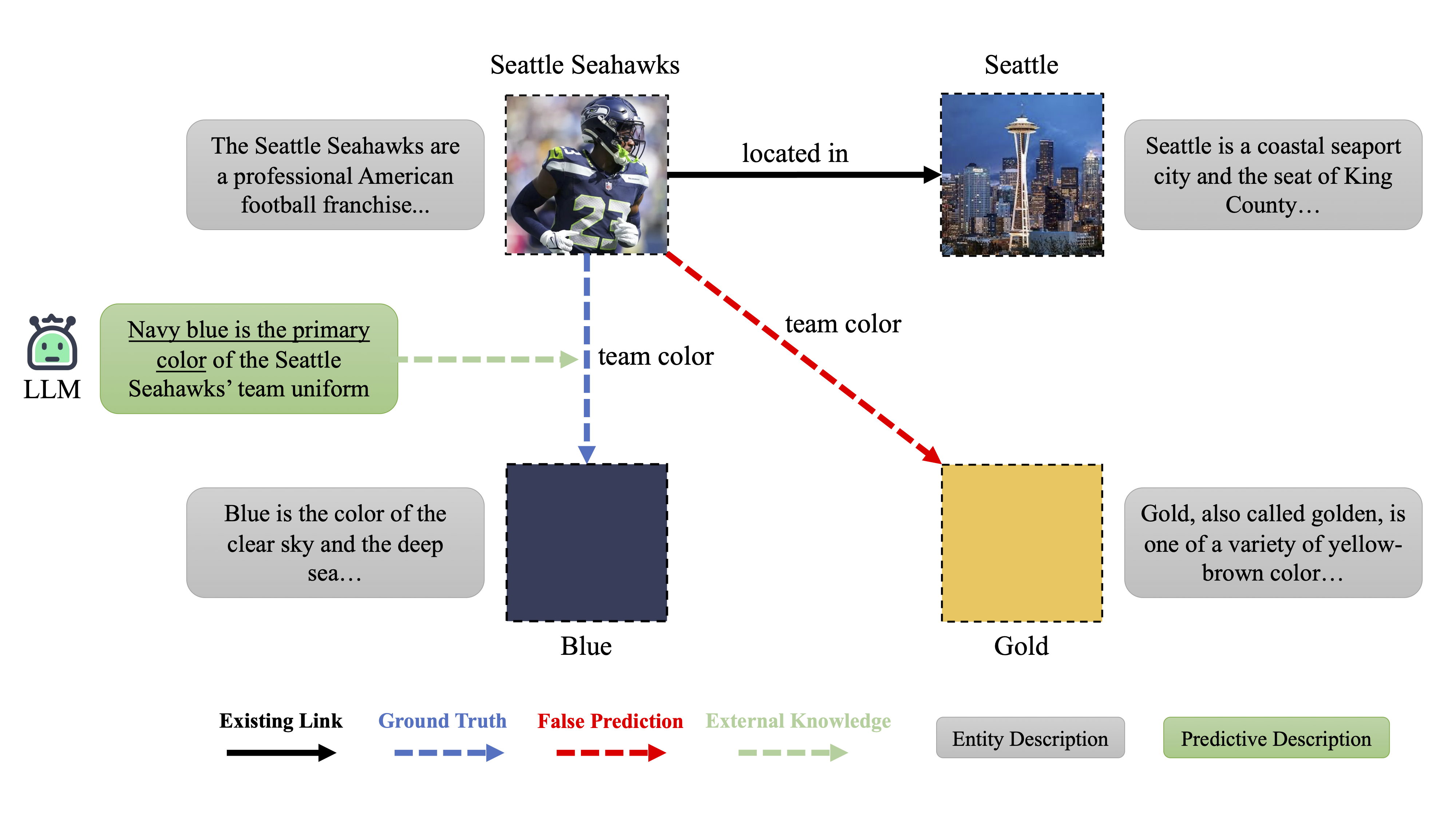}
  \caption{An example of KGC from the FB15k-237 dataset in a text-based setting. Entities associated with their descriptions are connected by relational links.}
  \label{fig:kg_example}
\end{figure*}

To address the KGC challenge, text-based methods \cite{yao2019kg, wang2021structure, wang-etal-2022-simkgc} have become increasingly sophisticated, with notable approaches representing triples through textual descriptions of the involved entities and relations. These methods leverage pre-trained language models (PLMs) for fine-tuning, encoding queries and answers as separate continuous vectors. A standout in this domain is SimKGC \cite{wang-etal-2022-simkgc}, which employs a bi-encoder contrastive learning framework to achieve state-of-the-art performance. Despite their advancements, most text-based KGC methods face two significant limitations due to data inconsistency: \emph{mismatched descriptions} and \emph{pseudo-inverses}.

The \emph{mismatched description} issue arises because entity and relation descriptions are collected independently, leading to insufficient information for the query to predict the correct entity. For instance, in the FB15k-237 dataset \cite{dettmers2018convolutional}, as shown in Figure \ref{fig:kg_example}, the description of the head entity "Seattle Seahawks" lacks information about "team color". Consequently, a text-based model might struggle to choose between "Blue" and "Gold" when presented with the query $(\emph{Seattle Seahawks}, \emph{team color}, ?)$. This issue can be alleviated by obtaining a predictive description based on the query from an external knowledge base to improve descriptive accuracy. The \emph{pseudo-inverse} problem arises from simplistic data augmentation techniques used in current methods. For example, in SimKGC, an inverse relation $r^{-1}$ is created by simply adding the prefix "reverse", resulting in two queries, $(h,r, ?)$ and $(t,r^{-1}, ?)$, for each triple $(h,r,t)$. This approach can confuse the model, making it difficult to accurately identify the inverse relation $r^{-1}$, potentially leading to overfitting issues. For example, semantically, the inverse "similar to" is more straightforward and explicit than using "reverse similar to" for the relation "similar to".
 
In addition, text-based methods often employ contrastive loss in a self-supervised manner, overlooking the label information inherent in a KG. In this paper, we introduce \textbf{K}nowledge graph completion of \textbf{E}nhanced \textbf{R}elation \textbf{M}odeling with \textbf{I}nverse \textbf{T}ransformation (KERMIT) to address these challenges. To resolve the issue of "mismatched description", we propose leveraging large language models (LLMs) as an external knowledge source to generate semantically coherent descriptions for KGC queries. Subsequently, we enhance the existing KG by meticulously crafting inverse relations, ensuring symmetry and coherence between queries and answers. It is important to emphasize that our method, as a fundamental data-level approach, can be applied to the majority of text-based methods, significantly improving their performance on KGC tasks. Lastly, we consider entities that satisfy the same query $(h,r,?)$ to belong to the same class, thereby enabling fully supervised contrastive learning \cite{khosla2020supervised} to obtain more expressive representations. Empirical experiments validate the efficacy of our method, achieving state-of-the-art performance on the WN18RR \cite{toutanova2015representing} and FB15k-237 \cite{dettmers2018convolutional} datasets. In summary, our main contributions are as follows:

\begin{itemize}
    \item We propose to utilize LLMs to automatically address the "mismatched descriptions" and "pseudo-inverse" issues prevalent in text-based methods, with detailed discussions on prompting techniques and model selection.
    \item We further propose to assign class labels to candidate answer entities from the underlying KG and extend the contrastive learning framework used by text-based methods to accommodate multiple positives per query.
    \item We conduct experiments on benchmark datasets with comprehensive baseline comparison to evaluate the KGC performance of our approach. Furthermore, we perform an extensive analysis of KERMIT to demonstrate its effectiveness.
\end{itemize}

%% file: related_work.tex
\section{Related work}
\textbf{Knowledge graph completion}. The task of KGC has traditionally centered on designing effective scoring mechanisms to assess the plausibility of triples in low-dimensional embedding spaces. A foundational approach in this area is knowledge graph embedding (KGE), commonly known as embedding-based methods. Representative techniques include TransE \citep{bordes2013translating}, DistMult \citep{yang2014embedding}, RotatE \citep{sun2018rotate}, ConvE \citep{dettmers2018convolutional}, and TuckER \citep{balazevic-etal-2019-tucker}. Despite their success in KGC performance, these approaches are often opaque, functioning as black boxes and making them difficult for humans to understand. Alternatively, rule-based methods such as NeuralLP \citep{yang2017differentiable}, DRUM \citep{sadeghian2019drum}, RNNLogic \citep{qu2020rnnlogic}, and LERP \citep{han2023logical} focus on learning first-order logical Horn clauses for KGC. Despite their clear interpretability, rule-based methods often struggle with scalability to large graphs and suffer from inefficiencies in training.

\textbf{Text-based KGC methods}. Recent developments in KGC have emphasized scoring a triple $(h, r, t)$ using textual information from the entities and relations involved. DKRL \citep{xie2016representation} leverages a deep convolutional neural network (CNN) to encode entity descriptions in a bag-of-words format. MTL-KGC \citep{kim-etal-2020-multi} adopts a multi-task learning framework for text-based KGC tasks. Meanwhile, KG-BERT \citep{yao2019kg} and StAR \citep{wang2021structure} utilize two separate pre-trained language models with identical initial weights to encode queries and answers. SimKGC \citep{wang-etal-2022-simkgc} introduces a contrastive learning framework \citep{pmlr-v119-chen20j} combined with advanced negative sampling methods, which is further improved by GHN \citep{qiao2023improving} through the generation of semantic-enhanced negatives, resulting in notable advancements for text-based KGC. Recently, large language models (LLMs) have emerged as powerful tools, leading to methods like CP-KGC \citep{zhang2023making} and KICGPT \citep{wei2024kicgpt} that leverage LLMs to advance KGC.

\textbf{Prompt engineering of LLMs}. Effective interaction with LLMs increasingly relies on prompt engineering \citep{marvin2023prompt}. Role-play prompting, a strategy where LLMs are assigned specific expert roles, has been shown to enhance domain-specific performance \citep{kong2024self, shao2023character}. Furthermore, LLMs exhibit exceptional in-context learning capabilities, handling various downstream tasks by conditioning on prompts consisting of a few illustrative input-output examples \citep{ma2024fairness}. Encouraging LLMs to generate additional explanations improves both the interpretability of their outputs and their overall accuracy \citep{jie2024interpretable}. In this work, we employ these techniques to develop a tailored prompting template that effectively addresses challenges such as "mismatched descriptions" and "pseudo-inverse".

\textbf{Supervised contrastive learning}. Contrastive learning has become a cornerstone of unsupervised learning, significantly advancing representation learning by encouraging augmented views of the same data point to be similar, while pushing apart representations of different data points \citep{technologies9010002, benshaul2023reverse}. Foundational works like SimCLR \cite{pmlr-v119-chen20j} and MoCo \citep{he2020momentum} have validated its potential for learning effective visual representations without requiring labels. At its heart lies the contrastive loss, such as InfoNCE \cite{pmlr-v119-chen20j, ijcai2019p746}, which promotes learning features that are both informative and discriminative. The work of \cite{khosla2020supervised} extends this concept to a fully supervised batch contrastive setting, effectively leveraging label information by accommodating multiple positives per anchor. More recently, contrastive learning has been successfully extended to other domains, such as natural language processing \cite{zhao2023leveraging, sidiropoulos2024improving}, illustrating its adaptability and potential. Despite these advances, its application to graph data remains underexplored.

%% file: preliminaries.tex
\section{Preliminaries}\label{sec:preliminaries}

\textbf{Knowledge graph completion}. We model a knowledge graph (KG) as $\mathcal{G} \! = \! (\mathcal{E}, \mathcal{R}, \mathcal{T}, \mathcal{D})$, where $\mathcal{E}$ and $\mathcal{R}$ denote the sets of entities and binary relations, respectively. $\mathcal{T} \! = \! \{ (h,r,t) | h,t \! \in \! \mathcal{E}, r \! \in \! \mathcal{R} \}$ represents the set of factual triples, each forming a directed labeled edge $h \mathop{\rightarrow}\limits^{r} t$ in this KG. $\mathcal{D}$ maps each entity or relation to its textual description. For example, $\mathcal{D}(\text{"Seattle Seahawks"}) = \text{"The Seattle Seahawks are a..."}$ provides the description for the entity "Seattle Seahawks". Given a query $(h,r,?)$, the task of knowledge graph completion (KGC) is to identify the missing tail entity by retrieving the most plausible candidate $\hat{t}$ from the entity set $\mathcal{E}$, satisfying $(h,r,\hat{t})$.

\textbf{Text-based methods for KGC}. To ensure this paper is self-contained, we will provide a general introduction to the principles of text-based methods. Typically, a text-based model is designed to learn a score function $f(\mathcal{D}(h), \mathcal{D}(r), \mathcal{D}(t))$ for a specific triple $(h,r,t)$ based on its textual description, where a higher score indicates greater plausibility. For simplicity, we denote the score function as $f(h,r,t)$. The majority of text-based approaches employ BERT \cite{devlin-etal-2019-bert} to encode the textual information of a triple $(h,r,t)$ into semantic space. Two independently functioning BERT encoders with the same initial pre-trained weights are used to process the input $(h, r)$ and the output $t$. The embedding of the $(h, r)$ pair is noted as $e_{hr}$, while $e_t$ denotes the embedding of the tail entity. We will now examine the steps for generating $e_{hr}$ and $e_t$ embeddings:

Leaving $H_\text{desc}, R_\text{desc}, T_\text{desc}$ be the tokenized textual descriptions of $h, r, t$, we have
\begin{align}\label{eq:bert_encode}
C_{hr} & = \text{BERT}_\text{hr}(X_{hr}), ~~ C_{t} = \text{BERT}_\text{t}(X_{t})
\end{align}
where $X_{hr}$ and $X_{t}$ are concatenated sequences of tokens:
\begin{align}
X_{hr} & = \left[ x_\text{CLS}, H_\text{desc}, x_\text{SEP}, R_\text{desc}, x_\text{SEP} \right] \label{eq:hr_token_sequence} \\
X_{t} & = \left[ x_\text{CLS}, T_\text{desc}, x_\text{SEP} \right] \label{eq:t_token_sequence}
\end{align}
where $x_\text{CLS}$ and $x_\text{SEP}$ are special tokens introduced by \cite{devlin-etal-2019-bert}. $C_{hr}$ and $C_{t}$ are the corresponding representation vectors.

Then, a pooling strategy is employed to aggregate a sequence of representations into a final embedding vector:
\begin{align}\label{eq:pool}
e_{hr} & = \text{Pool}(C_{hr}), ~~ e_{t} = \text{Pool}(C_{t})
\end{align}

In common practice, CLS pooling is used in downstream NLP tasks, utilizing the CLS token's embedding vector to represent the whole sentence. The similarity score is then calculated using the cosine function between the two embeddings:
\begin{equation}
    \cos(e_{hr}, e_t) = \frac{e_{hr} \cdot e_{t}}{||e_{hr}|| \cdot ||e_{t}||}
\end{equation}

Finally, following the idea of contrastive learning \cite{technologies9010002, benshaul2023reverse} for acquiring expressive representations, InfoNCE loss, augmented with an additive margin \cite{pmlr-v119-chen20j, ijcai2019p746}, is employed to construct the training objective:
\begin{equation}\label{eq:contrastive_loss}
\mathcal{L} = -\log \frac{e^{(f(h,r,t)-\gamma) / \tau}}{e^{(f(h,r,t)-\gamma) / \tau} + \sum_{t' \in \mathcal{N}^-}{e^{(f(h,r,t')-\gamma) / \tau}}}
\end{equation}
where $\mathcal{N}^-$ is the entity set of negative samples, $\gamma$ the additive margin,  $\tau$ the temperature and $f(h,r,t) = \cos(e_{hr}, e_t) \in [-1,1]$ the score of the triple $(h,r,t)$.

During inference, we first compute the embeddings of all entities $t \in \mathcal{E}$ ahead of time. Then, for a missing triple $(h,r,?)$ from the testing set, a ranking list of all candidate entities is generated by measuring the similarity between $e_{hr}$ and the entity embeddings.
\begin{equation}
    \hat{t} = \mathop{\arg\max}\limits_{t} \cos(e_{hr}, e_t), t \in \mathcal{E}
\end{equation}

\textbf{Prompting LLMs}. Basically, we utilize LLMs to retrieve responses with the prompting technique \cite{liu2023pre} in a question-and-answer process. To do this, the initial step involves the design of a prompting template denoted as \textbf{P}, which incorporates a placeholder \{X\} intended for accommodating the input $\mathbf{x}$. Subsequently, a specific input $\mathbf{x}$ is inserted into this placeholder, thereby creating the prompt that serves as the direct input to the LLMs. Finally, we collect the generated output $\mathbf{y}$ produced by the LLMs. This procedural sequence can be formally articulated as follows:
\begin{equation}
    \mathbf{y} = \text{LLM}(\mathbf{x} | \textbf{P})
\end{equation}

%% file: methodology.tex
\input{prompting_template_pred_desc}

\section{Methodology}

This section first proposes our method, namely "predictive description", designed to tackle the challenges of "mismatched description" and "pseudo-inverse" with the aid of LLMs. Following this, we will provide an detailed explanation of our model's architecture and training strategy, incorporating a supervised contrastive loss built upon the inherent label information of a KG.

\subsection{Predictive description via LLMs}
Under the formulation in Section \ref{sec:preliminaries}, the KGC task here is considered to be an entity prediction problem \cite{malaviya2020commonsense}. From this point of view, the textual description plays an important role in featuring an entity.

Although encoding a triple through its textual information yields good performance for KGC tasks, relying solely on descriptions of the head entity and relation usually fails to accurately predict the tail. The issue arises when the description of a query $(h, r)$ lacks coherence with the tail $t$, resulting in suboptimal predictions. Thereby, we propose to use LLMs, as they can serve as an implicit database for vast open-domain knowledge \cite{shao2023prompting, li2022self}, to generate predictive information aimed at the tail entity with a well-designed prompting template $\textbf{P}_\text{desc}$:
\begin{equation}\label{eq:t_pred}
    t_\text{pred} = \text{LLM}(g_k(h, r) | \textbf{P}_\text{desc})
\end{equation}

\input{prompting_template_inv_rel}

The prompting template we use, as part of the input to LLMs, is shown in Table \ref{tab:prompting_template_pred_desc}. The function $g_k(\cdot)$ is a pre-processing step that fetches textual descriptions and retrieves $k$ few-shot triple examples sharing the same relation $r$. We fix the number of few-shot examples $k$ to $5$ during all experiments. In this prompt, we utilize role-playing \cite{kong2024self, shao2023character} and few-shot engineering techniques \cite{ma2024fairness} to instruct the LLM to generate a list of potential tail entities that could complete the query $(h,r,?)$, along with general descriptions of these entities. Additionally, an explanation from the LLM is requested to ensure accuracy \cite{jie2024interpretable}. Finally, we employ simple concatenation to obtain the predictive description of the query: "\{entities\}: \{description\}".

For head entity prediction, we replace $h$ and $r$ with $t$ and $r'$ in Equation \ref{eq:t_pred}, where $r'$ is the inverse of $r$, resulting in:
\begin{equation}\label{eq:h_pred}
    h_\text{pred} = \text{LLM}(g_k(t, r') | \textbf{P}_\text{desc})
\end{equation}

LLMs offer a cost-effective and efficient mechanism for obtaining textual information, obviating the necessity for human intervention. We also notice that some recent methods solve entity prediction problems by involving LLMs as external knowledge engines, e.g. Prophet \cite{shao2023prompting}, a knowledge-based visual question answering (VQA) model, teaches GPT-3 to answer visual questions from given candidates by showing examples selected by a vanilla VQA model to it, which is a certain case of in-context learning \cite{dong2022survey}. The task of KGC might also be handled through the combination of a vanilla KGC model and an LLM. We leave the investigation of this idea for future work.

Addressing the challenge of "pseudo-inverse" in KGC, it is important to note that each node can act both as a head and a tail. Considering this duality, we propose converting the relations to their inverse forms to maintain semantic coherence. Similarly, we utilize LLMs to identify these inverse relations, employing the prompting template displayed in Table \ref{tab:prompting_template_inv_rel}:
\begin{equation}\label{eq:inv_rel}
    r' = \text{LLM}(g_k(r) | \textbf{P}_\text{rel})
\end{equation}

Here, the prompt $\textbf{P}_\text{rel}$ utilizes analogous prompting techniques as $\textbf{P}_\text{desc}$ in Table \ref{tab:prompting_template_pred_desc}. The inverse relation can then be straightforwardly extracted from the LLMs' response.

\begin{figure*}[t]
  \centering
  \includegraphics[width=.9\textwidth]{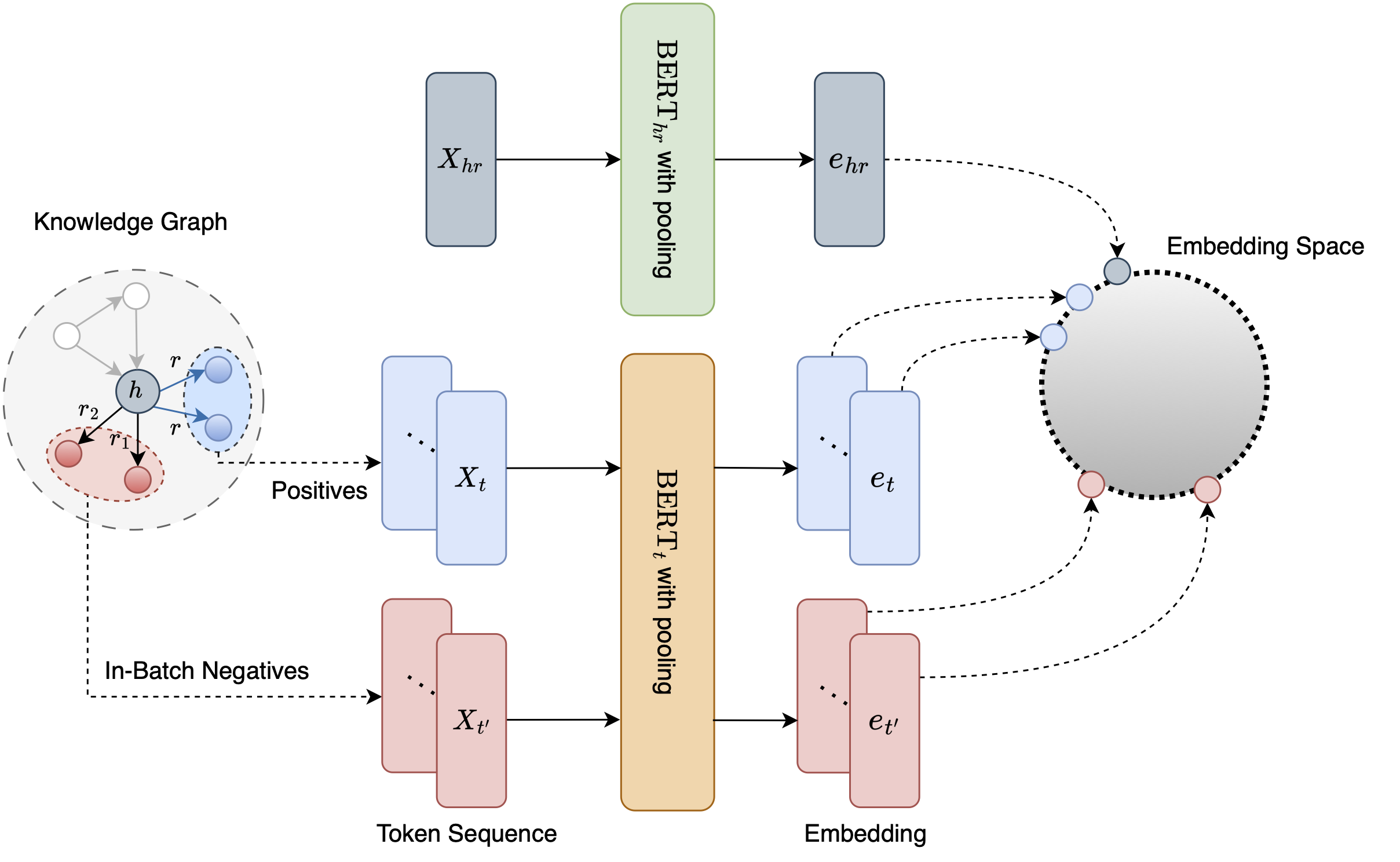}
  \caption{An overview of KERMIT during training. For a given query $(h,r,?)$, KERMIT begins by sampling a subset of positive entities $\{t | t \in \mathcal{E}, (h,r,t) \in \mathcal{T}\}$ and a subset of negative entities $\{t' | t' \in \mathcal{E}, (h,r,t') \notin \mathcal{T}\}$. The query and potential answer entities are then converted into token sequences, followed by two separate BERT encoders to generate embeddings. Finally, a supervised contrastive loss is applied to optimize the encoders.}
  \label{fig:model_architecture}
\end{figure*}

\subsection{Model architecture}
In order to take full advantage of the generated predictive descriptions, we propose a novel method to tokenize the text sequence. Let $H_\text{desc}, R_\text{desc}, T_\text{desc}, T_\text{pred}$ be the corresponding token sequences of the descriptions of $h, r, t$ and the predictive information $t_\text{pred}$, we separate $H_\text{desc}, R_\text{desc}, T_\text{pred}$ by the special token $x_\text{SEP}$ as $X_{hr}$, while $X_t$ remains the same as Equation \ref{eq:t_token_sequence}.
\begin{align}\label{eq:tokenize_sep}
X_{hr} = [ x_\text{CLS}, H_\text{desc}, x_\text{SEP}, R_\text{desc}, x_\text{SEP}, T_\text{pred}, x_\text{SEP} ]
\end{align}

Following SimKGC \cite{wang-etal-2022-simkgc}, we utilize two BERT \cite{devlin-etal-2019-bert} encoders with identical pre-trained parameters, operating independently during training and inference for different representation learning purposes.

Given a triple $(h, r, t)$, the encoded vectors of $(h,r)$, and $t$, denoted as $e_{hr}$ and $e_{t}$, are computed following Equation \ref{eq:bert_encode} and \ref{eq:pool}. For the pooling method, we diverge from the commonly used CLS token and instead employ mean pooling, which has been shown to produce superior sentence embeddings \citep{reimers-gurevych-2019-sentence}. An overview of our model is shown in Figure \ref{fig:model_architecture}.

\subsection{Supervised contrastive training}
In order to enhance the expressiveness of query and entity representations, we propose to effectively leverage the label information inherent in a KG, thus formulating a supervised variant \cite{khosla2020supervised} of the self-supervised contrastive loss shown in Equation \ref{eq:contrastive_loss}:
\begin{equation}\label{eq:contrastive_loss_sup}
\mathcal{L} = - \frac{1}{|\mathcal{N}^+|} \sum_{t \in \mathcal{N}^+} \log \frac{e^{(f(h,r,t)-\gamma) / \tau}}{e^{(f(h,r,t)-\gamma) / \tau} + \sum_{t' \in \mathcal{N}^-}{e^{(f(h,r,t')-\gamma) / \tau}}}
\end{equation}
where $\mathcal{N}^+$ denotes the set of positive entities for a specific query $(h,r,?)$, encompassing tail entities from triples with the same head $h$ and relation $r$. This formulation allows for multiple positive samples per query, ensuring more expressive and generalizable outcomes.

For sampling positive and negative entities, we utilize an in-batch strategy \cite{khosla2020supervised, pmlr-v119-chen20j} for the sake of efficiency. In this approach, class labels are assigned to entities within the same batch to indicate their positivity or negativity. Exploring more effective sampling strategies is reserved for future work. For instance, an entity's embedding can be calculated using its textual description through a pre-trained language model, and entities with the lowest semantic similarity can be considered negatives.

Note that the predictive descriptions (\emph{i.e}. $t_\text{pred}$ and $h_\text{pred}$) and inverse relations for all training and testing samples are obtained within a number of chat completion API invocations in advance, which is not part of the training and inference flow.

%% file: prompting_template_pred_desc.tex
\begin{table}[t]
    \caption{The prompting template used to generate the predictive description of the tail entity includes placeholders for the head entity's name (h\_name) and its textual description (h\_desc), the relation name (r\_name), and a few-shot example of triples (triple\_examples). These placeholders will be replaced when dealing with an incomplete triple $(h, r, ?)$.}
    \centering
    \small
    \noindent\fbox{%
    \begin{minipage}{\linewidth} 
\tt
You are a learned knowledge creator. I have an entity "\{h\_name\}" and a relation "\{r\_name\}".
Here is the description of the entity:\\
\{h\_name\}: \{h\_desc\}\\

There are several examples for you in which two entities satisfy the relation "\{r\_name\}". Each example is in the form of a triple (h, r, t), meaning that the relationship between the entities "h" and "t" is "r". For instance, the fact that "Bob is the father of Tom" can be expressed by the triple (Bob, father of, Tom). Each example takes one single line:\\
\{triple\_examples\}\\

Your task is to find an entity such that the relationship between "\{h\_name\}" and the entity is "\{r\_name\}". First, list at least five potential entities that might satisfy the requirement. Then, generate a broad description of the common properties shared by the entities. Your response must follow the JSON format below: \\
\{\\
    "entities": [entity1, entity2, ...],\\
    "description": "a broad description of the common properties shared by the entities",\\
    "explanation": "explanation"\\
\}\\
The description is limited to 30 words and no less than 10 words. The explanation should detail why you choose the entities based on the given entity "\{h\_name\}", relation "\{r\_name\}" and triple examples.
    \end{minipage}
}
    \label{tab:prompting_template_pred_desc}
\end{table}

%% file: prompting_template_inv_rel.tex
\begin{table}[t]
    \caption{The prompting template used to generate inverse relations includes placeholders the relation name (r\_name), and a few-shot example of triples (triple\_examples). These placeholders will be replaced when dealing with a specific relation $r$.}
    \centering
    \small
    \noindent\fbox{%
    \begin{minipage}{\linewidth} 
\tt
You are a learned knowledge creator. I have a relation "\{r\_name\}". There are several examples for you in which two entities satisfy the relation "\{r\_name\}". Each example is in the form of a triple (h, r, t), meaning that the relationship between the entities "h" and "t" is "r". For instance, the fact that "Bob is the father of Tom" can be expressed by the triple (Bob, father of, Tom). Each example takes one single line:\\
\{triple\_examples\}\\

Your task is to find an inverse (r') of the relation "\{r\_name\}" such that the form (t, r', h) holds true for all the given triples. Generate a potential relation that might satisfy the requirement. Your response must follow the JSON format below: \\
\{\\
    "inverse relation": "the inverse of relation '\{r\_name\}'",\\
    "explanation": "explanation"\\
\}\\
The explanation should detail why you choose the inverse based on the given relation "\{r\_name\}" and triple examples.
    \end{minipage}
}
    \label{tab:prompting_template_inv_rel}
\end{table}

%% file: experiments.tex
\input{data_statistics}

\section{Experiments}
\subsection{Experiment setting}\label{subsec:exp_setting}

\textbf{Data preparation}. To evaluate our methods for completing missing triples in KGs, we select three datasets for the KGC task: UMLS \cite{kok2007statistical}, WN18RR and FB15k-237 \cite{wang2021kepler}. The statistics of each dataset are presented in Table \ref{tab:data-statistics}. The UMLS comprises biomedical concepts, including drug and disease names, as well as relations between them, such as diagnosis and treatment. The WN18 and FB15k datasets were initially introduced in the work by \cite{bordes2013translating}. Subsequent research \cite{dettmers2018convolutional, toutanova2015representing}, however, revealed that these two datasets are susceptible to test leakage. This issue led to the creation of refined versions: WN18RR and FB15k-237. The textual descriptions of WN18RR and FB15k-237 are sourced from the data made available by KG-BERT \cite{yao2019kg} and CP-KGC \cite{yang2024enhancing}, respectively.

During the training process for the task of KGC, each triple $(h,r,t)$ is augmented by creating two queries: $(h,r,?)$ and $(t,r',?)$, with the corresponding answers $t$ and $h$. Here, $r'$ represents the inverse of $r$. Since obtaining $h_\text{pred}$ in Equation \ref{eq:h_pred} necessitates inverse relations, we utilize \emph{gpt4-turbo-2024-04-09} to generate these inverses, as outlined in Equation \ref{eq:inv_rel}. The pairs of $r$ and $r'$ for the WN18RR dataset are listed in Table \ref{tab:relationships_wn18rr}.

However, dealing with the FB15k-237 dataset appears to be more complicated. In contrast to the WN18RR dataset, the relations found within FB15k-237 pose a challenge due to their intractability for direct input into a text-based model. To address this concern, we undertake sentence formation of the relations and identify their corresponding inverses with the aid of an LLM. Given the substantial number (237) of relations within the FB15k-237 dataset, we opt to present only a limited subset of these relations in Table \ref{tab:relationships_fb15k237}.

\textbf{Evaluation Metrics}. For each query $(h, r, ?)$ or $(t,r',?)$, our approach computes a score for each entity, along with determining the rank of the correct answer. The Mean Reciprocal Rank (MRR) and Hit@$k$ values under the filtered protocol \cite{bordes2013translating} are reported based on the computed ranks across all queries. MRR calculates the mean of the reciprocal ranks for the answer entities, while Hit@$k$ determines the percentage of desired entities ranked within the top $k$ positions. The filtered protocol excludes the scores of all known true triples in the training, validation, and testing sets.

\textbf{Implementation details}. To obtain a comprehensive evaluation on the selection of LLMs, we utilize various open-source LLMs to generate predictive descriptions, including \emph{Qwen2-7B-Instruct} \cite{qwen2}, \emph{phi-3-small-128k-instruct} \cite{abdin2024phi} and \emph{phi-3-medium-4k-instruct} \cite{abdin2024phi}. In this paper, we refer to these three LLMs as \emph{qwen2-7b}, \emph{phi3-small} and \emph{phi3-medium} for simplicity. A subset of triple examples of WN18RR with textual descriptions and predictive descriptions via \emph{phi3-medium} is shown in Table \ref{tab:examples_wn18rr} (\marktext{more examples are shown in Table \ref{tab:examples_fb15k237} and Table \ref{tab:examples_wn18rr_llms}}). For additional information, please see Appendix \ref{app:experiment_settings}.

\input{experiment_results}

\subsection{Main results}
In this section, we report empirical results of our method, along with those of the chosen algorithms. We adopt the numerical results for NeuralLP, DRUM, RNNLogic, and ConvE as presented in \cite{qu2020rnnlogic}. For LERP, we refer to the values provided in \cite{han2023logical}. The metrics for TransE, DistMult, RotatE, TuckER, KG-BERT, MTL-KGC, StAR, and SimKGC are sourced from \cite{wang-etal-2022-simkgc, zhang2023making}.

As shown in Table \ref{tab:experiment_results}, our proposed model achieves state-of-the-art performance on the WN18RR and UMLS dataset, surpassing all other methods across all evaluation metrics. Compared to the baseline SimKGC on WN18RR, our model significantly improves MRR (from 66.6\% to 70.0\%), Hit@1 (from 58.7\% to 62.9\%), and other key metrics. Similar results can also be observed on UMLS. On the FB15k-237 dataset, KERMIT outperforms other approaches except for KICGPT \citep{wei2024kicgpt}. The success of KICGPT on FB15k-237 may attribute to the massive demonstrations required by in-context learning used by KICGPT, for athe FB15k-237 dataset exbihits a much denser topological structure.

As presented in Table \ref{tab:experiment_results}, our proposed model delivers state-of-the-art performance on the WN18RR and UMLS datasets, outperforming all competing methods across all evaluation metrics. On WN18RR, our model significantly surpasses our baseline, SimKGC, achieving an MRR improvement from 66.6\% to 70.0\% and a Hit@1 increase from 58.7\% to 62.9\%, alongside other metric enhancements. The UMLS dataset exhibits similar trends, further validating our model’s effectiveness. On the FB15k-237 dataset, KERMIT demonstrates strong performance, outperforming all methods except for KICGPT \citep{wei2024kicgpt}. The superior results of KICGPT on FB15k-237 may be attributed to its reliance on extensive demonstrations in its in-context learning strategy, particularly suited for the dataset’s much denser topological structure.

Additionally, we present the results obtained using our method with different LLM options to generate predictive descriptions, as shown in Table \ref{tab:experiment_results}. As expected, the most capable model, \emph{phi3-medium}, demonstrates the best or nearly best performance. Nevertheless, the differences in evaluation metrics among the selected LLMs are minimal, highlighting the effectiveness and robustness of our approach in choosing LLMs.

\input{ablation_pred_desp}

\begin{figure*}[t]
  \centering
  \includegraphics[width=1\textwidth]{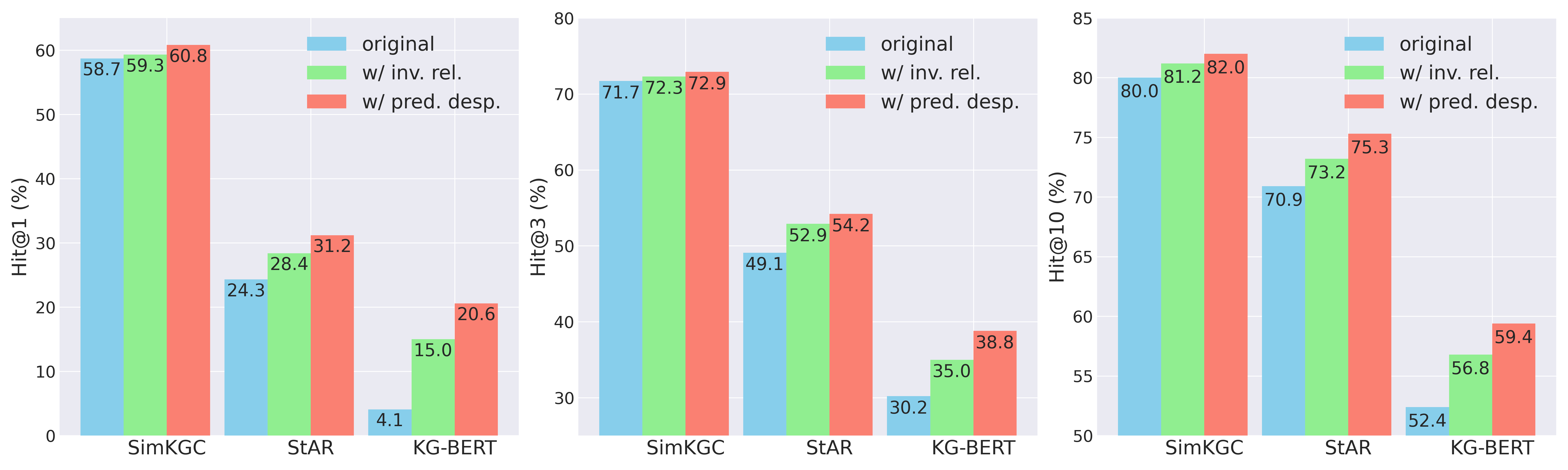}
  \caption{Model performance of text-based methods with the proposed approaches on WN18RR dataset. "w/ inv. rel.": with only inverse relations; "w/ pred. desc": with predictive descriptions. The predictive descriptions are collected using \emph{phi3-medium}.}
  \label{fig:ablation_pred_desc_wn18rr}
\end{figure*}

\subsection{Ablation results}
A key contribution of our work is the use of LLMs to incorporate external knowledge and identify inverse relations, which significantly enhances KGC target predictions. Additionally, extracting class labels inherent in KGs to develop a multi-positive contrastive loss is crucial for effective representation learning. We evaluate the impact of these components by successively removing them and assessing the performance on WN18RR and FB15k-237 datasets, with results detailed in Table \ref{tab:ablation_pred_desp}.

On the WN18RR dataset, utilizing supervised contrastive loss significantly improves MRR by 1.0\% and other metrics by a clear margin, as shown in Table\ref{tab:ablation_pred_desp} (c) \emph{vs}. (d). While Hit@1 increases from 61.0\% to 62.2\% due to the introduction of few-shot triple examples in LLMs' prompts, the metrics for Hit@3 and Hit@10 are adversely affected, Table\ref{tab:ablation_pred_desp} (b) \emph{vs}. (d). We hypothesize that while few-shot examples provide LLMs with additional knowledge, they may also restrict the creativity of LLMs, resulting in and poorer generalization of our KGC model. Fortunately, the supervised contrastive loss mitigates over-fitting dilemma through multiple positive samples, thereby enhancing overall metrics, Table\ref{tab:ablation_pred_desp} (a). Similar observations can be made on the FB15k-237 dataset, with the exception that the overall performance decreases due to few-shot examples (e.g., MRR decreases by 0.6\%), as shown in Table \ref{tab:ablation_pred_desp} (b) \emph{vs}. (d). This limited impact may be attributed to the dataset's prevalent 1-to-many relations and its denser topological structure compared to WN18RR (more edges and fewer nodes), challenging LLMs' accurate entity prediction, as evidenced by multiple tail candidates like "protein" and "iron" for the query (\emph{beef}, \emph{food has nutrient}, ?) (see Table \ref{tab:relationships_fb15k237}). The potential of combining textual information with the graph's topological structure in our approach is a promising direction for future research.

To further demonstrate the effectiveness and generalization ability of predictive descriptions and curated inverse relations, we apply our method to various text-based models, including SimKGC \cite{wang-etal-2022-simkgc}, StAR \cite{wang2021structure}, and KG-BERT \cite{yao2019kg}, evaluating their performance on the WN18RR and FB15k-237 datasets (see Figure \ref{fig:ablation_pred_desc_wn18rr} and \ref{fig:ablation_pred_desc_fb15k237}). The results show a distinct improvement across all metrics after incorporating predictive descriptions and inverse relations, demonstrating our method's effectiveness as a general enhancement tool for text-based models.

%% file: data_statistics.tex
\begin{table}[h]
  \caption{Statistics of datasets.}
  \centering
  \resizebox{\linewidth}{!}
  {\begin{tabular}{cccccc}
    \toprule
    Dataset & \# Relation & \# Entity & \# Train & \# Validation & \# Test \\
    \midrule
    UMLS & 46 & 135 & 5327 & 596 & 633 \\
    WN18RR & 11 & 40,943 & 86,835 & 3,034 & 3,134 \\
    FB15k-237 & 237 & 14,541 & 272,115 & 17,533 & 20,466 \\
    \bottomrule
  \end{tabular}}
  \label{tab:data-statistics}
\end{table}

%% file: experiment_results.tex
\begin{table*}[t]
  \caption{
    Knowledge graph completion results for and WN18RR and FB15k-237 datasets. All metrics are in \%. The best performance for each metric within each dataset is highlighted in bold, and the best metrics among categories are underlined.
    }
  \centering
  \resizebox{\linewidth}{!}{\begin{tabular}{ccccccccccccc}
    \toprule
    & \multicolumn{4}{c}{WN18RR} & \multicolumn{4}{c}{FB15k-237} & \multicolumn{4}{c}{UMLS} \\
    \cmidrule(lr){2-5} \cmidrule(lr){6-9} \cmidrule(lr){10-13}
    & MRR & Hit@1 & Hit@3 & Hit@10 & MRR & Hit@1 & Hit@3 & Hit@10 & MRR & Hit@1 & Hit@3 & Hit@10 \\
    \midrule
    NeuralLP \cite{yang2017differentiable} & 38.1 & 36.8 & 38.6 & 40.8 & 23.7 & 17.3 & 25.9 & 36.1 & 48.3 & 33.2 & 56.3 & 77.5 \\ 
    DRUM \cite{sadeghian2019drum} & 38.2 & 36.9 & 38.8 & 41.0 & 23.8 &  17.4 &  26.1 & 36.4 & 69.5 & 54.6 & 80.8 & 93.5 \\
    RNNLogic \cite{qu2020rnnlogic} & 51.3 & 47.1 & 53.2 & 59.7 & \underline{34.9} & \underline{25.8} & \underline{38.5} & \underline{53.3} & 74.5 & 63.0 & 83.3 & 92.4 \\
    LERP \cite{han2023logical} & \underline{62.2} & \underline{59.3} & \underline{63.4} & \underline{68.2} & - & - & - & - & \underline{76.2} & \underline{64.4} & \underline{85.5} & \underline{94.2} \\
    \midrule
    TransE \cite{bordes2013translating} & 24.3 & 4.3 & 44.1 & 53.2 & 27.9 & 19.8 & 37.6 & 44.1 & - & - & - & 98.9 \\
    DistMult \cite{yang2014embedding} & 44.4 & 41.2 & 47.0 & 50.4 & 28.1 & 19.9 & 30.1 & 44.6 & 39.1 & 25.6 & 44.5 & 66.9 \\
    RotatE \cite{sun2018rotate} & \underline{47.6} & 42.8 & \underline{49.2} & \underline{57.1} & 33.8 & 24.1 & 37.5 & 53.3 & \underline{74.4} & \underline{63.6} & \underline{82.2} & 93.9 \\
    ConvE \cite{dettmers2018convolutional} & 43.0 & 40.0 & 44.0 & 52.0 & 32.5 & 23.7 & 35.6 & 50.1 & - & - & - & \underline{99.0} \\
    TuckER \cite{balazevic-etal-2019-tucker} & 47.0 & \underline{44.3} & 48.2 & 52.6 & \underline{35.8} & \underline{26.6} & \underline{39.4} & \underline{54.4} & 73.2 & 62.5 & 81.2 & 90.0 \\
    \midrule
    KG-BERT \cite{yao2019kg} & 21.6 & 4.1 & 30.2 & 52.4 & - & - & - & 42.0 & - & - & - & \underline{99.0} \\
    MTL-KGC \cite{kim-etal-2020-multi} & 33.1 & 20.3 & 38.3 & 59.7 & 26.7 & 17.2 & 29.8 & 45.8 & - & - & - & - \\
    StAR \cite{wang2021structure} & 40.1 & 24.3 & 49.1 & 70.9 & 29.6 & 20.5 & 32.2 & 48.2 & - & - & - & - \\
    SimKGC \cite{wang-etal-2022-simkgc} & 66.6 & 58.7 & 71.7 & 80.0 & 33.6 & 24.9 & 36.2 & 51.1 & \underline{79.4} & \underline{70.4} & \underline{86.5} & 94.4 \\
    GHN \cite{qiao2023improving} & \underline{67.8} & 59.6 & 71.9 & \underline{82.1} & 33.9 & 25.1 & 36.4 & 51.8 & - & - & - & - \\
    CP-KGC \cite{zhang2023making} & 67.3 & \underline{59.9} & \underline{72.1} & 80.4 & 33.8 & 25.1 & 36.5 & 51.6 & - & - & - & - \\
    \marktext{KICGPT} \cite{wei2024kicgpt} & 54.9 & 47.4 & 58.5 & 64.1 & \textbf{41.2} & \textbf{32.7} & \textbf{44.8} & \textbf{55.4} & - & - & - & - \\
    \midrule
    \rowcolor{GrayBG} KERMIT (qwen2-7b) & 69.4 & 62.0 & 73.7 & \textbf{83.3} & 35.1 & 25.8 & 38.7 & 53.7 & 91.6 & 84.5 & \textbf{98.6} & \textbf{99.5} \\
    \rowcolor{GrayBG} KERMIT (phi3-small) & 69.5 & 62.1 & 73.7 & 83.0 & 35.2 & 25.8 & 38.8 & 53.9 & \textbf{92.1} & \textbf{85.6} & 98.3 & \textbf{99.5} \\
    \rowcolor{GrayBG} KERMIT (phi3-medium) & \textbf{70.0} & \textbf{62.9} & \textbf{73.8} & 83.2 & \underline{35.9} & \underline{26.6} & \underline{39.6} & \underline{54.7} & 91.9 & 85.2 & \textbf{98.6} & 99.4 \\
    \bottomrule
  \end{tabular}}
  \label{tab:experiment_results}
\end{table*}

%% file: ablation_pred_desp.tex
\begin{table*}[t]
  \caption{
    Knowledge graph completion accuracy of KERMIT and its variants on WN18RR and FB15k-237. "pred-desc": with predictive descriptions; "few-shot": with few-shot triple examples when generating predictive descriptions; "sup": with supervised contrastive loss.
  }
  \centering
  \resizebox{\linewidth}{!}{\begin{tabular}{c|ccc|cccc|cccc}
    \toprule
    & & & & \multicolumn{4}{c|}{WN18RR} & \multicolumn{4}{c}{FB15k-237} \\
    KERMIT (phi3-medium) & pred-desc & few-shot & sup & MRR & Hit@1 & Hit@3 & Hit@10 & MRR & Hit@1 & Hit@3 & Hit@10 \\
    \midrule
    \rowcolor{GrayBG} (a) & \checkmark & \checkmark & \checkmark & \textbf{70.0} & \textbf{62.9} & \textbf{73.8} & 83.2 & \textbf{35.9} & \textbf{26.6} & \textbf{39.6} & \textbf{54.7} \\
    (b) & \checkmark & \checkmark & & 69.3 & 62.2 & 73.2 & 82.5 & 34.1 & 25.0 & 37.3 & 52.5 \\
    (c) & \checkmark & & \checkmark & 69.6 & 62.4 & 73.7 & \textbf{83.5} & 35.2 & 25.7 & 38.8 & 53.9 \\
    (d) & \checkmark & & & 68.6 & 61.0 & 73.1 & 82.9 & 34.7 & 25.1 & 37.4 & 52.5 \\
    \bottomrule
  \end{tabular}}
  \label{tab:ablation_pred_desp}
\end{table*}

%% file: analysis.tex
\begin{figure*}[t]
  \centering
  \includegraphics[width=.75\textwidth]{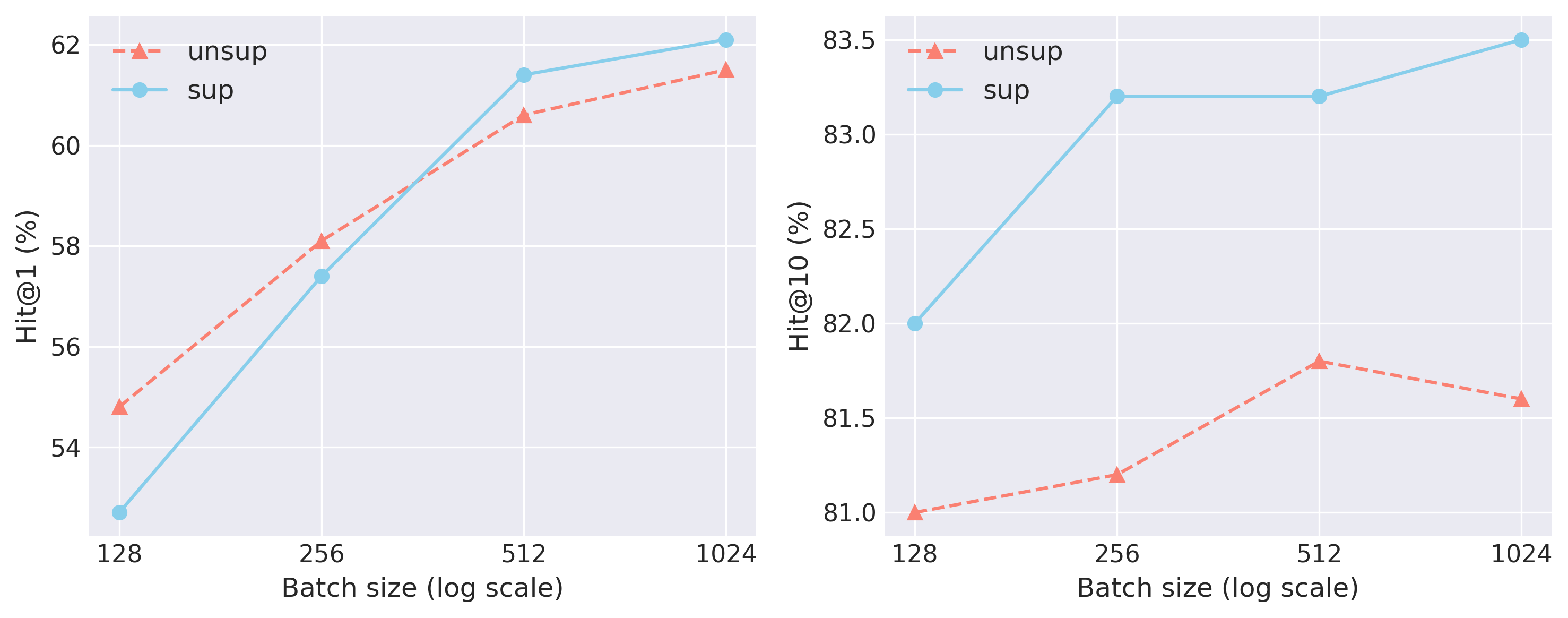}
  \caption{Performance on the WN18RR dataset \emph{w.r.t.} batch size as achieved by KERMIT. "sup" indicates training with the supervised contrastive strategy, whereas "unsup" represents training with the unsupervised approach.}
  \label{fig:ablation_sup_contras}
\end{figure*}

\input{analysis_tokenization}
\input{analysis_fewshot}

\section{Analysis}
\subsection{Fine-grained analysis}
\textbf{Supervised contrastive loss}. In Figure \ref{fig:ablation_sup_contras}, we present a quantitative analysis of how Hit@1 and Hit@10 are affected by the increasing batch size. When the batch size is relatively small, the scarcity of positive samples causes supervised contrastive training to underperform unsupervised training in terms of Hit@1 metrics. As the batch size grows to 512, however, supervised training reliably exceeds unsupervised performance. This finding underscores the necessity of sufficient positive samples for supervised contrastive training to achieve its optimal accuracy. Furthermore, the markedly superior Hit@10 results for supervised training highlight its robust generalization ability.

\textbf{Tokenization}. The introduction of transformer encoders leads to significant performance variations in text-based methods depending on the tokenization strategy used. We explore various heuristic methods for integrating predictive descriptions into queries, with evaluation results presented in Table \ref{tab:analysis_tokenization}. In our experiment, we use predictive descriptions generated by \emph{phi3-medium} while maintaining the hyperparameters unchanged.

The first strategy, as shown in Equation \ref{eq:tokenize_sep}, employs the 
special token $x_\text{SEP}$ for sentence separation. Another strategy, differing slightly in its implementation, uses a SPACE token $x_\text{SPACE}$:
\begin{equation}\label{eq:tokenize_space}
X_{hr} = [x_\text{CLS}, H_\text{desc}, x_\text{SEP}, R_\text{desc}, x_\text{SPACE}, T_\text{pred}, x_\text{SEP} ]
\end{equation}

However, since the operation complexity of multi-head attention \cite{vaswani2017attention} grows quadratically as the length of the input sequence increases \cite{shazeer2019fast}, extra information might cause unexpected computational overhead. An alternative version is to represent $X_{hr}$ with just the $T_\text{pred}$ to reduce the computational cost, resulting in Equation \ref{eq:tokenize_only_pred}.
\begin{equation}\label{eq:tokenize_only_pred}
    X'_{hr} = \left[ x_\text{CLS}, T_\text{pred}, x_\text{SEP} \right]
\end{equation}

\textbf{Few-shot triple examples} play an important role in the prompting template we use to instruct LLMs, as illustrated in Table \ref{tab:prompting_template_pred_desc}. We examine the impact of these few-shot examples by comparing the predictive descriptions generated by \emph{phi3-medium} with and without their inclusion in the prompt, as shown in Table \ref{tab:analysis_fewshot}. When presented with the query (\emph{cover}, \emph{derivationally related form}, ?), incorporating few-shot examples clearly enhances the LLM's understanding of the task. This results in a more precise and focused predictive description over the unknown answer "covering".

\subsection{Case study on selected LLMs}
The capabilities of LLMs typically correspond to their parameter sizes and training datasets. Generating predictive descriptions is a complex task for LLMs, requiring skills like semantic understanding of tasks, knowledge extraction, and text summarization. In this section, we present the predictive descriptions produced by the three selected LLMs, as illustrated in Table \ref{tab:examples_wn18rr_llms}.

As anticipated, the most advanced of the three selected LLMs, \emph{phi3-medium}, produces more precise and informative descriptions for predicting the desired entities. Additionally, the bilingual LLM \emph{qwen2-7b} tends to struggle with complex English scenarios, leading to lower-quality descriptions compared to \emph{phi3-small}. However, the overall differences in generating predictive descriptions among these three LLMs are minimal, as evidenced by the close KGC results displayed in Table \ref{tab:experiment_results}.

%% file: analysis_tokenization.tex
\begin{table*}[t]
  \caption{
   Results given by different tokenization strategies of the query sequence on the WN18RR and FB15k-237 datasets. "sep", "space" and "only $T_\text{pred}$" refer to the tokenization strategies specified by Equation \ref{eq:tokenize_sep}, \ref{eq:tokenize_space} and \ref{eq:tokenize_only_pred}, respectively.
  }
  \centering
  \begin{tabular}{ccccccccc}
    \toprule
    & \multicolumn{4}{c}{WN18RR} & \multicolumn{4}{c}{FB15k-237} \\
    \cmidrule(lr){2-5} \cmidrule(lr){6-9}
    & MRR & Hit@1 & Hit@3 & Hit@10 & MRR & Hit@1 & Hit@3 & Hit@10 \\
    \midrule
    \rowcolor{GrayBG} sep & \textbf{70.0} & \textbf{62.9} & \textbf{73.8} & \textbf{83.2} & \textbf{35.9} & \textbf{26.6} & \textbf{39.6} & \textbf{54.7} \\
    space & 68.2 & 60.6 & 71.5 & 81.6 & 34.4 & 24.9 & 37.0 & 52.0 \\
    only $T_\text{pred}$ & 48.0 & 37.4 & 54.9 & 68.1 & 28.6 & 22.0 & 31.5 & 43.0 \\
    \bottomrule
  \end{tabular}
  \label{tab:analysis_tokenization}
\end{table*}

%% file: analysis_fewshot.tex
\begin{table*}[t]
  \caption{Predictive descriptions of the triple (\emph{cover}, \emph{derivationally related form}, \emph{covering}) generated by \emph{phi3-medium}, both with and without few-shot examples in the prompt. Descriptions of the entities involved are available in Table \ref{tab:examples_wn18rr}.}
  \centering
  \begin{tabularx}{\textwidth}{lX}
    \toprule
    case & Predictive Description ($t_\text{pred}$) \\
    \midrule
    w/ few-shot & \underline{covering}, conceal, protect, shield, enclose: these entities are actions or objects that are associated with \underline{the act of covering}, hiding, or protecting something. \\
    w/o few-shot & uncover, coverage, coverlet, recover, discover: these entities are derived from the root 'cover', sharing similar core meanings but varying slightly in usage and context. \\
    \bottomrule
  \end{tabularx}
  \label{tab:analysis_fewshot}
\end{table*}

%% file: conclusion.tex
\section{Conclusion}
In this paper, we introduce KERMIT to address two major challenges in text-based knowledge graph completion (KGC): the "mismatched description" and "pseudo-inverse" issues. Our approach encompasses two main strategies: employing large language models (LLMs) for generating predictive descriptions and curating inverse relations. Additionally, we develop a supervised contrastive loss that allows for multiple positive samples to enhance KGC performance. Experimental results demonstrate that KERMIT significantly outperforms baseline methods and is adaptable across different text-based approaches. Future work will explore the use of LLMs for KGC, with a focus on prompting techniques such as Chain-of-Thought (CoT).

%% file: broader_impact.tex
\section*{Broader impacts}
Data quality and its match with the specific task are key factors in determining the performance of machine learning models. To the best of our knowledge, we are the first to pinpoint key data-level issues that hinder the performance of text-based KGC models. Our method significantly enhances a variety of text-based approaches, including but not limited to SimKGC and KG-BERT. By dealing with the root cause of the issue, we anticipate that our work will steer the KGC community towards recognizing the challenges imposed by data characteristics, emphasizing the critical need for data-focused solutions to enhance model performance.

%% file: acknowledgements.tex
\section*{Acknowledgements}
This work is supported by the Shandong Key Research and Development Program (No. 2023CXPT065), National Natural Science Foundation of China (No. 62272129) and Taishan Scholars (No. tsqn202408112).

%% file: appendix.tex
\section{Experiment settings}\label{app:experiment_settings}
\textbf{Comparison of algorithms}. In experiment, the performance of our model is compared with that of the following algorithms: (1) Rule learning methods. We have selected a set of algorithms that specialize in learning logical rules for the purpose of completing missing triples. This selection allows for a comprehensive comparison, and the chosen algorithms are NeuralLP \cite{yang2017differentiable}, DRUM \cite{sadeghian2019drum}, RNNLogic \cite{qu2020rnnlogic} and LERP \cite{han2023logical}. (2) Embedding-based methods. Furthermore, we extend our comparison to include various embedding methods, such as TransE \cite{bordes2013translating}, DistMult \cite{yang2014embedding}, RotatE \cite{sun2018rotate}, ConvE \cite{dettmers2018convolutional} and TuckER \cite{balazevic-etal-2019-tucker}. These models adopt an approach where representations for entities and relations are learned respectively. (3) Text-based methods. We select the following models that also leverage textual information into our consideration: KG-BERT \cite{yao2019kg}, MTL-KGC \cite{kim-etal-2020-multi}, StAR \cite{wang2021structure}, SimKGC \cite{wang-etal-2022-simkgc}, GHN \cite{qiao2023improving}, CP-KGC \cite{zhang2023making} and KICGPT \cite{wei2024kicgpt}.

\textbf{Additional implementation details}. To ensure a fair comparison with previous studies, we initialize the BERT encoders using pretrained weights from \emph{bert-base-uncased} (English). The AdamW optimizer \cite{loshchilov2018decoupled} with linear learning rate decay is employed to optimize the model parameters. The initial learning rates are set to $8 \times 10^{-5}$ for the WN18RR dataset and $1 \times 10^{-5}$ for the FB15k-237 dataset. We maintain a fixed batch size of 1024. The temperature parameter $\tau$ is tuned by searching within the set $\{ 0.02, 0.03, 0.04 \}$, based on results from 65 epochs for WN18RR and 10 epochs for FB15k-237.

\section{Additional experiment results}
\subsection{Additional ablation results}
\begin{figure*}[h]
  \centering
  \includegraphics[width=1\textwidth]{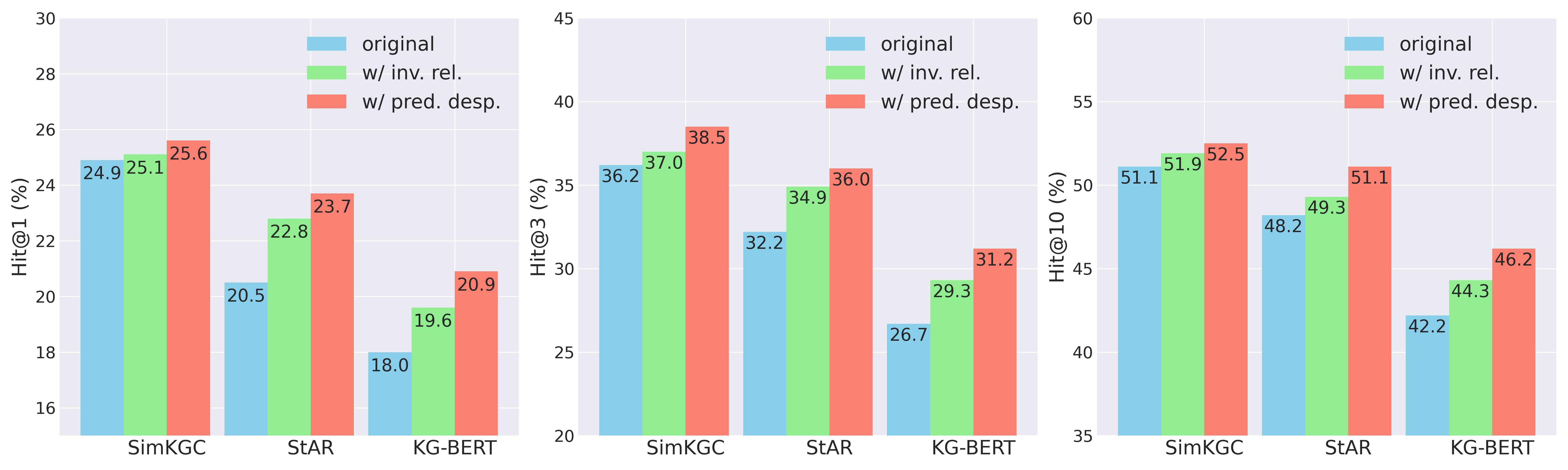}
  \caption{Model performance of text-based methods with the proposed approaches on FB15k-237 dataset. "w/ inv. rel.": with only inverse relations; "w/ pred. desc": with predictive descriptions. The predictive descriptions are collected using \emph{phi3-medium}.}
  \label{fig:ablation_pred_desc_fb15k237}
\end{figure*}

\subsection{Inverse relations}
Table \ref{tab:relationships_wn18rr} presents the relations and their inverses from the WN18RR dataset. For the FB15k-237 dataset, we show results in Table \ref{tab:relationships_fb15k237}.

\input{relationships_wn18rr}
\input{relationships_fb15k}
\clearpage

\subsection{Predictive descriptions}
Table \ref{tab:examples_wn18rr} and \ref{tab:examples_fb15k237} present additional examples of predictive descriptions for the WN18RR and FB15k-237 datasets, respectively.

\input{examples_wn18rr}
\input{examples_fb15k_more}
\clearpage

\section{Additional analysis results}
The results of further analysis on the selection of different LLMs are shown in Table \ref{tab:examples_wn18rr_llms}.
\input{examples_wn18rr_llms}

%% file: relationships_wn18rr.tex
\begin{table*}[H]
  \caption{Relations and their corresponding inverses within the WN18RR dataset.}
  \centering
  \begin{tabular}{ccc}
    \toprule
    Relation ($r$) & Inverse Relation ($r'$) & Example ($h, t$) \\
    \midrule
    hypernym & hyponym & \text{retionalize}, \text{think}  \\
    derivationally related form & derivationally related form & \text{decision making}, \text{settle} \\
    instance hypernym & instance hyponym & \text{Paterson}, \text{urban center} \\
    also see & also see & \text{write}, \text{write up} \\
    member meronym & whole holonym & \text{primulaceae}, \text{sea milkwort} \\
    synset domain topic of & has synset domain topic & \text{valency}, \text{biology} \\
    has part & is part of & \text{Europe}, \text{Republic of Estonia} \\
    member of domain usage & has characteristic or instance of & \text{colloquialism}, \text{firewater} \\
    member of domain region & located in & \text{Scotland}, \text{Scottish} \\
    verb group & verb group & \text{supplant}, \text{replace} \\
    similar to & similar to & \text{advantageous}, \text{meanwhile} \\
    \bottomrule
  \end{tabular}
  \label{tab:relationships_wn18rr}
\end{table*}

%% file: relationships_fb15k.tex
\newcommand{\raw}{{\greentext{Raw:}}}
\newcommand{\norm}{\newline{\greentext{Sentence-forming($r$):}}}
\newcommand{\inv}{\newline{\greentext{Inverse($r'$):}}}
\newcommand{\tripleexample}{\newline{\greentext{Examples($h, t$):}}}

\begin{table*}[ht]
  \caption{A subset of the raw, normalized and inverse relations within the FB15k-237 dataset.}
    \centering
    \small
    \noindent\fbox{%
    \begin{minipage}{\linewidth} 
\tt
\raw{} /people/person/gender
\norm{} person's gender
\inv{} gender of person
\tripleexample{} (Rick Wakeman, Male), (Kellie Martin, Female) \\ \\

\raw{} /people/person/places\_lived./people/place\_lived/location
\norm{} person lived in place
\inv{} place of person lived in
\tripleexample{} (Chritina Hendricks, Portland), (Penelope Ann Miller, New York City) \\ \\

\raw{} /film/actor/film./film/performance/film
\norm{} actor of film
\inv{} film has actor
\tripleexample{} (Donald Sutherland, Six Degrees of Separation), (Robert Vaughn, The Towering Inferno) \\ \\

\raw{} /location/country/form\_of\_government
\norm{} government's form
\inv{} form of government
\tripleexample{} (New Zealand, Unitary state), (Netherlands, Parliamentary system), (Kenya, Republic) \\ \\

\raw{} /tv/tv\_program/regular\_cast./tv/regular\_tv\_appearance/actor
\norm{} tv program's regular actor
\inv{} regular actor of tv program
\tripleexample{} (Disney's House of Mouse, Russi Taylor), (Michael Jackson: 30th Anniversary Celebration, Marlon Brando) \\ \\

\raw{} /tv/tv\_writer/tv\_programs./tv/tv\_program\_writer\_relationship/tv\_program
\norm{} writer of tv program
\inv{} tv program's writer
\tripleexample{} (John Lutz, Saturday Night Live), (Damon Lindelof, Lost), (David Benioff, Game of Thrones) \\ \\

\raw{} /ice\_hockey/hockey\_team/current\_roster./sports/sports\_team\_roster/position
\norm{} ice hockey sports team roster has position
\inv{} position in ice hockey sports team roster
\tripleexample{} (Minnesota Wild, Defenseman), (Boston Bruins, Centerman), (Nashville Predators, Goaltender) \\ \\

\raw{} /food/food/nutrients./food/nutrition\_fact/nutrient
\norm{} food has nutrient
\inv{} nutrient in food
\tripleexample{} (Onion, Ash), (Carrot, Phenylalanine), (Beef, Polyunsaturated fat)
    \end{minipage}
    }

  \label{tab:relationships_fb15k237}
\end{table*}

%% file: examples_wn18rr.tex
\newcommand{\head}{{\greentext{Head name:}}}
\newcommand{\hd}{\newline{\greentext{Head desc:}}}
\newcommand{\rel}{\newline{\greentext{Relation($r$):}}}
\newcommand{\tail}{\newline{\greentext{Tail name:}}}
\newcommand{\td}{\newline{\greentext{Tail desc:}}}
\newcommand{\hrd}{\newline{\greentext{$t_\text{pred}$:}}}
\newcommand{\trd}{\newline{\greentext{$h_\text{pred}$:}}}

\begin{table}[h]
  \caption{Triple examples of the WN18RR dataset along with generated predictive descriptions via \emph{phi3-medium}.}
    \centering
    \small
    \noindent\fbox{%
    \begin{minipage}{\linewidth} 
\tt
\head{} land reform
\hd{} a redistribution of agricultural land (especially by government action)
\rel{} hypernym
\inv{} is a type of
\tail{} reform
\td{} a change for the better as a result of correcting abuses; "justice was for sale before the reform of the law courts"
\hrd{} Agricultural policy, Government action, Land redistribution, Social \underline{reform}, Tenure \underline{reform}: Policies, actions, or \underline{reforms} related to the redistribution or management of land, often involving government intervention.
\trd{} Legal \underline{reform}, Economic \underline{reform}, Educational \underline{reform}, Political \underline{reform}, Healthcare \underline{reform}: These are all systematic changes aimed at improving various aspects of society or specific sectors. \\

\head{} cover
\hd{} provide with a covering or cause to be covered; "cover her face with a handkerchief"; "cover the child with a blanket"; "cover the grave with flowers"
\rel{} derivationally related form
\inv{} derivationally related form
\tail{} covering
\td{} an artifact that covers something else (usually to protect or shelter or conceal it)
\hrd{} \underline{covering}, conceal, protect, shield, enclose: These entities are actions or objects that are associated with the act of covering, hiding, or protecting something.
\trd{} coating, shroud, envelope, \underline{cover}, sheath: These entities all involve a protective or concealing function, usually by enveloping or surrounding an object. \\

\head{} kamet
\hd{} a mountain in the Himalayas in northern India (25,450 feet high)
\rel{} instance hypernym
\inv{} has instance
\tail{} mountain peak
\td{} the summit of a mountain
\hrd{} \underline{mountain}, hill, \underline{peak}, \underline{summit}, ridge: Natural elevated landforms, part of the Earth's topography, characterized by height, steepness, and prominence.
\trd{} \underline{Mount} Everest, K2, \underline{Mount} Kilimanjaro, \underline{Mount} McKinley, \underline{Mount} Fuji: These are all iconic \underline{mountain} peaks, each being the highest in their respective regions. \\

\head{} class arachnida
\hd{} a large class of arthropods including spiders and ticks and scorpions and daddy longlegs; have four pairs of walking legs and no wings pointedly referring to or concerning a person's individual personality or intimate affairs especially offensively; "unnecessarily personalized remarks".
\rel{} derivationally related form
\inv{} derivationally related form
\tail{} spidery
\td{} relating to or resembling a member of the class Arachnida
\hrd{} \underline{arachnid, arachnophobia, arachnology, arachnomancy, arachnoid}: Entities related to \underline{arachnids}, their study, fear, and characteristics of \underline{spider-like structures}.
\trd{} \underline{arachnid, spider}, spindly, spiderweb, spiderling: All entities are related to \underline{spiders}, exhibiting characteristics or forms derived from 'spidery'.
    \end{minipage}
    }
  \label{tab:examples_wn18rr}
\end{table}

%% file: examples_fb15k_more.tex
\begin{table*}[ht]
  \caption{Examples of the FB15k-237 dataset along with generated predictive descriptions via \emph{phi3-medium}.}
    \centering
    \small
    \noindent\fbox{%
    \begin{minipage}{\linewidth} 
\tt
\head{} Rick Wakeman
\hd{} Rick Wakeman is an English keyboard player and songwriter who gained fame as the keyboardist for the progressive rock band Yes and is also known for his solo albums, contributions to a BBC comedy series, and hosting a radio show on Planet Rock.
\rel{} person's gender
\inv{} gender of person
\tail{} Male
\td{} Male organisms produce sperm and can reproduce sexually by fertilizing a female ovum, but can also reproduce asexually in some cases, and have a Y chromosome which codes for testosterone production to develop male reproductive organs, though sex determination can vary across species.
\hrd{} Humanity, Mankind, People, Males, Gender, \underline{Male}: These entities represent the gender category to which Rick Wakeman belongs.
\trd{} Female, Non-binary, Genderqueer, Agender, Cisgender: These entities represent different gender identities in relation to the male gender. \\ \\

\head{} Christina Hendricks
\hd{} Christina Rene Hendricks is an American actress renowned for her roles in AMC's Mad Men, for which she received four Emmy nominations, and FOX's Firefly, who was named "the sexiest woman in the world" and "Best Looking American Woman" by Esquire magazine.
\rel{} person lived in place
\inv{} place of person lived in
\tail{} Portland
\td{} Portland is a populous city in Oregon known for its environmental consciousness, public transportation networks, and outdoor activities, and is often referred to as the "City of Roses" due to its climate and abundance of rose gardens.
\hrd{} New York City, New York, USA, Detroit, Michigan, USA, Santa Monica, California, USA, Los Angeles, California, USA, Chicago, Illinois, USA: Geographical locations where Christina Hendricks has resided during her lifetime.
\trd{} Neil Young, Mike Rowe, MacLeod Andrews, Gordon Smith, Amanda Beard: Famous individuals who have resided in Portland, Oregon, contributing to its cultural and professional landscape. \\ \\

\head{} Donald Sutherland
\hd{} Donald McNichol Sutherland is a longstanding Canadian actor known for his roles in war movies and a variety of other films, and is the father of actor Kiefer Sutherland.
\rel{} actor of film
\inv{} film has actor
\tail{} Six Degrees of Separation
\td{} "Six Degrees of Separation" is a 1993 American drama film, based on a Pulitzer-nominated play by John Guare about real-life con artist David Hampton, featuring an Academy-nominated performance by Stockard Channing and referencing two Kandinsky artworks symbolizing chaos and control.
\hrd{} M*A*S*H, Ordinary People, Kramer vs. Kramer, Don't Look Now, A Clockwork Orange: These are all films in which Donald Sutherland has played a significant role.
\trd{} David Mamet, John Cusack, Will Smith, Stockard Channing, Kip Pardue: These are actors who have appeared in films together with an actor from "Six Degrees of Separation".
    \end{minipage}
    }

  \label{tab:examples_fb15k237}
\end{table*}

%% file: examples_wn18rr_llms.tex
\newcommand{\hrdqw}{\newline{\greentext{$t_\text{pred}$(qwen2-7b):}}}
\newcommand{\hrdps}{\newline{\greentext{$t_\text{pred}$(phi3-small):}}}
\newcommand{\hrdpm}{\newline{\greentext{$t_\text{pred}$(phi3-medium):}}}
\newcommand{\trdqw}{\newline{\greentext{$h_\text{pred}$(qwen2-7b):}}}
\newcommand{\trdps}{\newline{\greentext{$h_\text{pred}$(phi3-small):}}}
\newcommand{\trdpm}{\newline{\greentext{$h_\text{pred}$(phi3-medium):}}}

\begin{table*}[h]
  \caption{Triple examples of the WN18RR dataset along with generated predictive descriptions via three different LLMs.}
    \centering
    \small
    \noindent\fbox{%
    \begin{minipage}{\linewidth} 
\tt
\head{} trade name
\hd{} a name given to a product or service
\rel{} member of domain usage
\inv{} has characteristic or instance of
\tail{} metharbital
\td{} anticonvulsant drug (trade name Gemonil) used in the treatment of epilepsy
\hrdqw{} Lipitor: a cholesterol-lowering \underline{drug} (trade name Lipitor), Zoloft: an \underline{antidepressant} (trade name Zoloft), Tylenol: a pain reliever (trade name Tylenol), Advil: an anti-inflammatory (trade name Advil), Pepcid: a heartburn \underline{medication} (trade name Pepcid): \underline{medicinal} products with trade names used in specific \underline{health domains}.
\hrdps{} orphenadrine, cookie sheet, jail delivery, array, zimmer frame: Products, utensils, concepts, or tools associated with specific trade names, used within their respective domains.
\hrdpm{} product, service, advertising campaign, packet of illegal \underline{drugs}, elastic adhesive bandage: Entities used as trade names for specific products or services, often associated with their unique characteristics or functions.
\trdqw{} epilepsy treatment, anticonvulsant properties, drug formulation, medication for neurological disorders, pharmacological action: entities related to medical or pharmacological contexts.
\trdps{} anticonvulsant drug, epilepsy treatment, prescription medication, neurological disorder management, medical use: These entities are associated with the application, treatment, or categorization of metharbital.
\trdpm{} anticonvulsant, \underline{trade name}, drug used in the treatment of epilepsy, \underline{trade name Gemonil}, antiepileptic: The entities share properties of being related to 'metharbital', its usage, and its \underline{trade name}. \\

\head{} telephone
\hd{} get or try to get into communication (with someone) by telephone; "I tried to call you all night"; "Take two aspirin and call me in the morning"
\rel{} verb group
\inv{} verb group
\tail{} call
\td{} send a message or attempt to reach someone by radio, phone, etc.; make a signal to in order to transmit a message; "Hawaii is calling!"; "A transmitter in Samoa was heard calling"
\hrdqw{} \underline{call}, contact, reach, connect, communicate: These entities are related to actions involving telephone usage.
\hrdps{} communicate, \underline{call}, chat, talk, speak: Entities that involve exchanging information or conversation through vocal means.
\hrdpm{} \underline{call}, dial, ring, answer, hang up: These entities represent actions related to the use of a telephone, forming a verb group.
\trdqw{} ring, announce, signal, broadcast, speak: verb groups related to communication or message transmission
\trdps{} message, transmission, radio signal, \underline{phone call}, broadcast: Entities related to the process of sending or attempting to reach someone by various means.
\trdpm{} \underline{phone}, radio, signal, message, transmitter: These entities are associated with the act of \underline{communication} or transmitting information, similar to the entity 'call'.
    \end{minipage}
    }
  \label{tab:examples_wn18rr_llms}
\end{table*}